# CORE: Comprehensive Ontological Relation Evaluation for Large Language Models[1]

**Satyam Dwivedi[±*], Sanjukta Ghosh[†], Shivam Dwivedi[†], Nishi Kumari[±]**
**Anil Thakur[†], Anurag Purushottam[±]**
**Deepak Alok[‡], Praveen Gatla[§], Manjuprasad B[††], Bipasha Patgiri[‡‡]**
[±]Vaikhari AI, Bangalore [†]IIT BHU, Varanasi [‡]IIT Delhi, Delhi
[§]BHU, Varanasi [††]GSSSIETW, Mysore [‡‡]Tezpur University, Assam
satyam@vaikhari.ai

## Abstract

Large Language Models (LLMs) perform well on many reasoning benchmarks, yet existing evaluations rarely assess their ability to distinguish between meaningful semantic relations and genuine unrelatedness. We introduce CORE (Comprehensive Ontological Relation Evaluation), a dataset of 225K multiple-choice questions spanning 74 disciplines, together with a general-domain open-source benchmark of 203 rigorously validated questions (Cohen's κ = 1.0) covering 24 semantic relation types with equal representation of unrelated pairs. A human baseline from 1,000+ participants achieves 92.6% accuracy (95.1% on unrelated pairs). In contrast, 29 state-of-the-art LLMs achieve 48.25–70.9% overall accuracy, with near-ceiling performance on related pairs (86.5–100%) but severe degradation on unrelated pairs (0-41.35%), despite assigning similar confidence (≈92-94%). Expected Calibration Error increases 2-4x on unrelated pairs, and a mean semantic collapse rate of 37.6% indicates systematic generation of spurious relations. On the CORE 225K MCQs dataset, accuracy further drops to approximately 2%, highlighting substantial challenges in domain-specific semantic reasoning. We identify unrelatedness reasoning as a critical, under-evaluated frontier for LLM evaluation and safety.

## 1 Introduction

Large Language Models (LLMs) have demonstrated strong performance on reasoning benchmarks. Contemporary models achieve over 90% accuracy on MMLU benchmark (Hendrycks et al., 2020) and notable performance on specialized reasoning tasks. However, existing evaluations emphasize models' ability to recognize semantic relations when they exist (Bisk et al., 2020; Hupkes et al., 2020), with limited systematic evaluation of negative examples. A complementary and equally important capability remains largely unmeasured: reliably identifying cases where no meaningful semantic relation exists between concepts.

This oversight has practical consequences. In clinical decision support, systems must distinguish genuine symptom-disease correlations from spurious associations created through confounding variables. In financial trading, models must differentiate real market patterns from spurious associations; the Knight Capital Group's $440 million loss in 2012 is frequently cited as an example of automated system failure. In legal reasoning, AI systems must recognize when cases lack meaningful precedent; ChatGPT's fabrication of non-existent case citations in Mata v. Avianca (2023) resulted in legal sanctions. In scientific research, systems must avoid proposing false causal mechanisms based on surface-level statistical associations.

Across these domains, failures manifest not as random errors but as systematically confident false reasoning about relationships that do not exist. This subtle failure mode, i.e., confident construction of spurious relational structures rather than factual hallucination (Berglund et al., 2024; Liu et al., 2022; Mirzadeh et al., 2025; Petroni et al., 2019), is harder to detect and more dangerous in practice.

[1] The CORE benchmark and associated resources are available at core.vaikhari.ai; data and code are hosted at Hugging Face and GitHub.

* Corresponding Author

We introduce CORE (Comprehensive Ontological Relation Evaluation), a large-scale dataset of 225K multiple-choice questions spanning 74 disciplines. From this dataset we open-source the CORE benchmark: a general-domain evaluation subset of 203 rigorously validated questions targeting 24 semantic relation types with explicit balance between relational and unrelated concept pairs. This benchmark enables systematic measurement of a capability that has not been explicitly evaluated in prior work.

## 2 Related Work

Classical work on sense relations (Nuzzolese et al., 2016; Turney, 2005) established comprehensive frameworks for categorizing relationships in language. Recent applications of analogy reasoning to LLMs (Webb et al., 2023) have evaluated whether models can solve analogy problems. However, prior work primarily evaluates a narrow subset of valid analogies, without exhaustively testing major semantic relation types or assessing models' ability to recognize invalid or absent relations.

Work on model calibration (Guo et al., 2017; Kadavath et al., 2022) has examined whether model confidence aligns with accuracy. Calibration studies have primarily focused on balanced datasets and knowledge retrieval rather than systematic evaluation of performance asymmetries specific to absence of structure . Recent research on understanding what LLMs learn (Rogers et al., 2020; Yi et al., 2022) has examined whether models learn linguistic structure (Petroni et al., 2019), but has not systematically tested models' ability to recognize when structure is absent.

LLM reasoning has been extensively studied through frameworks examining emergent abilities (Wei et al., 2022) and improved reasoning strategies (Phan et al., 2025; Wang et al., 2022; Yao et al., 2023). Recent benchmarks have begun to address the "unanswerable" problem. *SimpleQA* (Wei et al., 2024) evaluates short-form factuality and refusal rates, while *AbstentionBench* (Kirichenko et al., 2025) demonstrates that reasoning-heavy models often degrade in their ability to refuse invalid premises. However, these studies do not evaluate reasoning on unrelated concept pairs. Our work addresses this gap.

## 3 Dataset and Benchmark Design

### 3.1 Overview and Scale

CORE comprises 225K multiple-choice questions spanning 74 disciplines across STEM, Humanities, and Social Sciences. This large corpus was constructed to support multiple purposes: fine-tuning, instruction-tuning, and evaluation. To mitigate evaluation contamination and overfitting risks, different portions serve different purposes. From this corpus, we define the **CORE benchmark**, an **open-source evaluation set** consisting of **203 general-domain questions** reserved exclusively for benchmarking. The benchmark is further divided into two subsets:

– **Open subset**: **102 questions** released for public analysis and evaluation.
– **Blind subset**: **101 questions** withheld for internal analysis and validation.

This benchmark focuses on 24 semantic relation types selected based on comprehensive ontological frameworks (Jullien et al., 2023) and validated through semantic evaluation methodologies. We maintain a **near-balanced distribution** between questions with related pairs (103) and unrelated pairs (100). The questions are designed to evaluate fundamental relational reasoning without domain-specific knowledge requirements.

### 3.2 Question Format and Design

Each question follows the analogical reasoning format: a reference concept pair (A:B) and an incomplete target pair (C:?), with four completion options. Questions employ everyday vocabulary, enabling evaluation of general semantic reasoning rather than specialized knowledge, aligning with HELM (Liang et al., 2022).

For related questions, each includes an explicit correct answer instantiating the target semantic relation:

**Question:** Artist is to brush as carpenter is to_?
**Options:**
**A:** Space **B:** House
**C:** Hammer **D:** Music
**Correct:** C: Hammer
**Relation:** Agent-Instrument
**Explanation:** An artist uses brush as their tool. Similarly, a carpenter uses hammer as their tool.

For unrelated questions, the initial concept pair lacks meaningful semantic relation, making the completion task ill-posed:

**Question:** Chess is to math as paper is to_?
**Options:**
**A:** Glass **B:** Plastic
**C:** Broccoli **D:** Cloth
**Correct:** C: Broccoli, acknowledging no meaningful relation exists
**Explanation:** Chess is unrelated to math in this context, just as paper is unrelated to broccoli. The other options have connections to paper.

### 3.3 Relation Types

CORE benchmark evaluates 24 distinct semantic relation types: agent-instrument, antonymy (complementary, converse, gradable), cause-effect, class-instance, co-hyponymy, entailment, function-object, homonymy, hyponymy, incompatibility, material-object, meronymy, metonymy, near-synonymy, part-substance, place-event, polysemy, presupposition, synonymy, troponymy, whole-process-step, and unrelated pairs.

### 3.4 Human Validation and Baseline

Answers and explanations for the CORE benchmark were initially developed for 250 questions and validated through a three-pass expert review process. The final benchmark comprises the 203 questions for which perfect inter-annotator agreement was achieved (Cohen's κ = 1.0). Each question includes human-authored explanations of why the correct answer instantiates the target relationship. Our annotation and validation process follows best practices (Andreas et al., 2013; Williams et al., 2018) with perfect inter-annotator agreement ensuring ground truth reliability.

Subsequently, a human baseline was constructed using responses from over 1,000 participants in India, spanning undergraduate to postdoctoral education levels, who completed the benchmark under blind evaluation conditions. **Table 1** reports aggregated human performance metrics.

| Category | Accuracy | Balanced Accuracy | Mean Entropy |
|---|---|---|---|
| Overall | 92.6% | 90.1% | 0.45 |
| Related pairs | 90.2% | 89.9% | 0.58 |
| Unrelated pairs | 95.1% | 95.1% | 0.31 |

**Table 1**: Human Performance on the Benchmark

Human baseline demonstrates that unrelated pair recognition is **not** inherently difficult; humans achieve 95.1% accuracy on unrelated pairs.

## 4 Evaluation Methodology

### 4.1 Model Selection and Coverage

We evaluate 29 state-of-the-art LLMs with cutoff date January 22, 2026. Our evaluation covers models from all major developers including GPT series (Achiam et al., 2023; Brown et al., 2020), Llama family (Touvron et al., 2023), Claude models (Anthropic, 2022), and compute-optimal models (Hoffmann et al., 2022).

Models were selected to achieve comprehensive coverage across all major developers and represent the frontier of capability. See Table 2 below for Model Details.

| Developer | Models |
|---|---|
| **Amazon** | Nova-2-lite, Nove-premier |
| **Anthropic** | Claude-Opus-4.5, Claude-Sonnet-4.5, Claude-Haiku-4.5 |
| **DeepSeek** | DeepSeek-R1, DeepSeek-V3.2 |
| **Google** | Gemini-3-flash, Gemini-2.5-pro, Gemini-2.5-flash |
| **Meta** | Llama-4-scout, Llama-4-maverick, Llama-3.3-70b-instruct, Llama-3.1-8b-instruct |
| **Mistral** | Mistral-Large-2512, Mistral-Nemo |
| **OpenAI** | GPT-5.2, GPT-5-mini, GPT-4o |
| **Misc.** | Grok-4.1-fast, Jamba-large-1.7, Kimi-k2-thinking, Perplexity-Sonar, Qwen3-max, Sarvam-m |
| **ZAI** | GLM-4.7, GLM-4.7-flash, GLM-4.6, GLM-4.6v-flash |

**Table 2**: Models Selected for Benchmark

Models span diverse architectures, sizes (8B to 405B parameters), and training approaches (supervised learning, RLHF, reasoning-focused).

### 4.2 Evaluation Protocol

All models were evaluated using a uniform prompting and evaluation protocol to ensure fair comparison, following established practices for evaluating emergent abilities in LLMs (Wang et al., 2022; Wei et al., 2022):

**For proprietary** models **(API access):** Deterministic inference with recommended inference settings to eliminate sampling variability. Models receive standardized question format and select from four options, reporting confidence across options.

**For open-source** models**:** Local inference using default configurations and identical hardware specifications.

Check **Appendix A. Model Inference prompt** for the prompt used for model inference.

### 4.3 Metrics

We employ multiple complementary metrics to characterize model performance.

**Notation:** Let $Q$ be the set of $N$ evaluated questions, indexed by $i$. For each question $i$:

- $y_i$: The ground truth semantic relation.
- $\hat{y}_i$: The model's predicted relation.
- $c_i \in [0,1]$ : The model's confidence score assigned to $\hat{y}_i$.
- Let $R$ be the set of all unique semantic relation types.
- $r_i$ : The specific relation type (e.g., agent–instrument, antonymy).
- $g_i \in \{\text{related,unrelated}\}$ : The high-level grouping of the pair.
- $\mathbb{1}[\cdot]$: The indicator function, evaluating to 1 if true and 0 otherwise.
- $B_1, \ldots, B_{10}$ : Disjoint confidence bins partitioning the predictions.

We define "correctness" as $1[\hat{y}_i = y_i]$ . Additionally, for error analysis, we define $h_i$ as a hallucination flag where $h_i = 1$ if ( $g_i = \text{unrelated} \wedge \hat{y}_i \neq \text{unrelated}$), and 0 otherwise.

Accuracy measures proportion of questions answered correctly:

$$\text{Accuracy} = \frac{1}{N}\sum_{i=1}^{N} 1[\hat{y}_i = y_i] \qquad (1)$$

**Balanced Accuracy** averages accuracy independently across semantic relation types:

$$\text{Balanced Accuracy} = \frac{1}{|R|}\sum_{r\in R} \text{Accuracy}_r \qquad (2)$$

**Expected Calibration Error (ECE)** measures systematic mismatch between predicted confidence and empirical accuracy across confidence bins:

$$\text{ECE} = \sum_{b=1}^{10} \frac{|B_b|}{N} |\text{acc}(B_b) - \text{conf}(B_b)| \qquad (3)$$

**Overconfident Error Rate** measures proportion of incorrect predictions with high confidence (≥0.75):

$$\text{OER} = \frac{1}{N}\sum_{i=1} 1[\hat{y}_i \neq y_i \wedge c_i \geq 0.75] \qquad (4)$$

**Semantic Collapse Rate** measures fraction of unrelated pairs misclassified as having semantic relation:

$$\text{SCR} = \frac{\sum_{i=1}^{N} h_i}{\sum_{i=1}^{N} \mathbb{1}[g_i = \text{unrelated}]} \qquad (5)$$

These metrics collectively characterize correctness, confidence calibration, high-stakes failure modes, and the specific failure mechanism of false relationship generation.

## 5 Results

### 5.1 Overall Performance and Asymmetry

Our empirical evaluation reveals patterns consistent with findings on other reasoning benchmarks (Clark et al., 2018; Mialon et al., 2023) but with a critical distinction. Overall accuracy ranges from ≈48.25–70.9% across 29 models. However, this masks a dramatic bifurcation:

| Category | Accuracy Range | Model Confidence Range |
|---|---|---|
| Overall | ≈48.25-70.90% | ≈92-95% |
| Related pairs | ≈86.50-100% | ≈93–95% |
| Unrelated pairs | ≈0-41.35% | ≈91–94% |

**Table 3**: Accuracy of models on the Benchmark

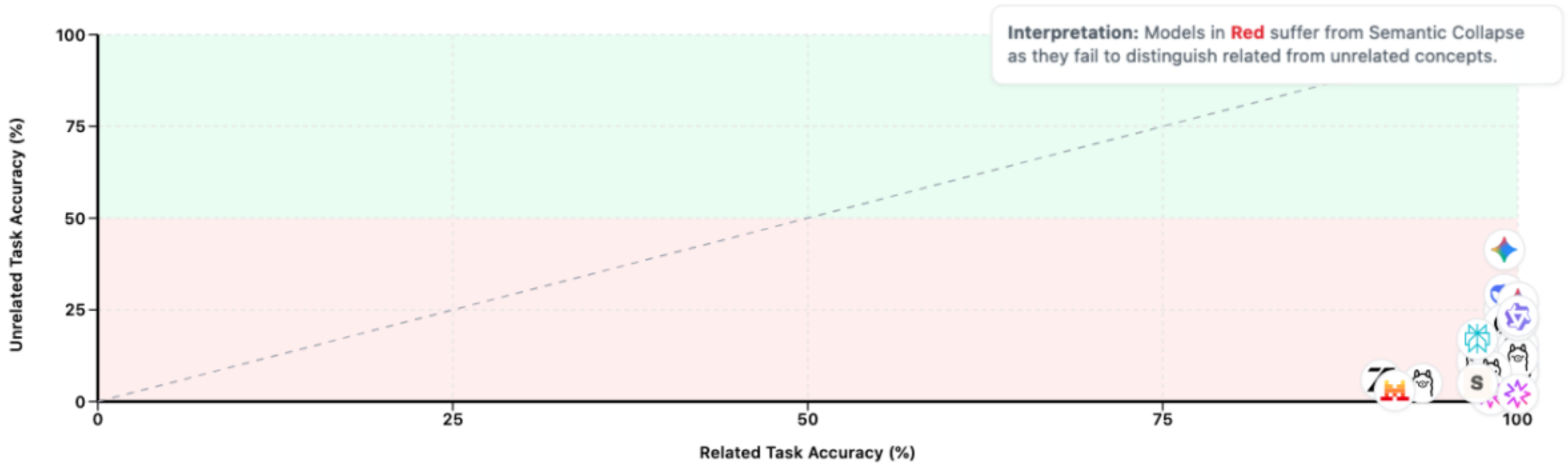

**Figure 1**: A comparison of model accuracy on tasks with related and unrelated pairs**.**

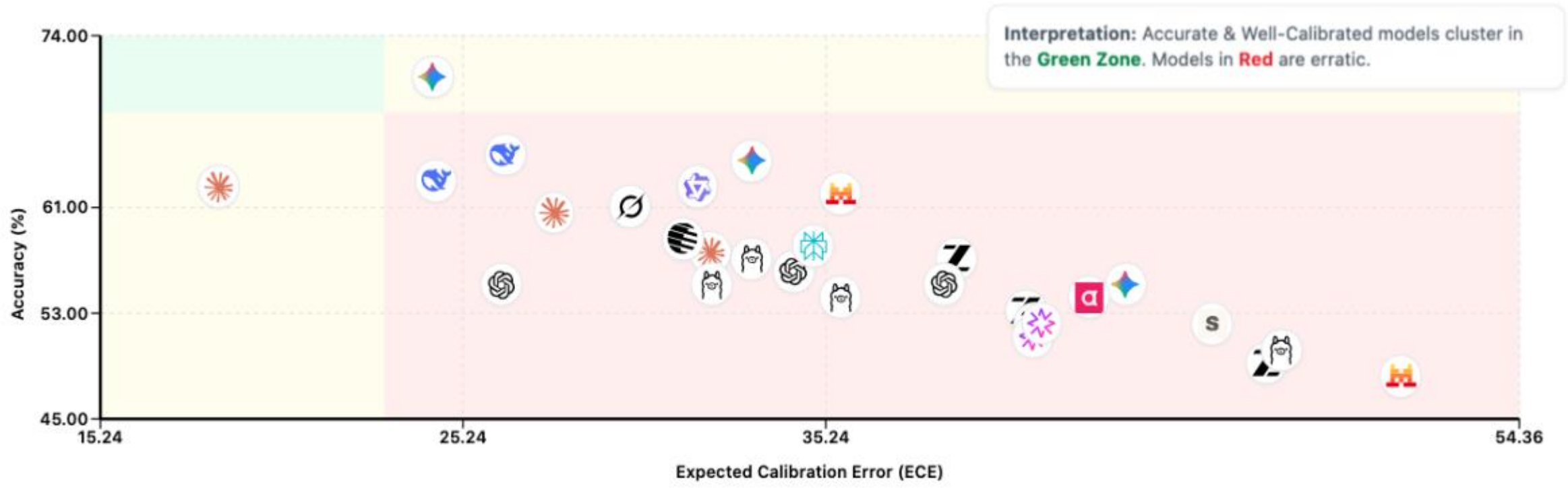

**Figure 2**: A comparison of models across Accuracy and Expected Calibration Error**.**

Despite 40–80 percentage point accuracy differences, models give near identical confidence on both related and unrelated tasks (related: ≈93–95%, unrelated: ≈91–94%), as illustrated in **Figure 1**. This **confidence–accuracy inversion** undermines the utility of confidence as a reliability signal, preventing downstream decision systems from appropriately weighting model outputs.

## 5.2 Calibration Analysis

**Expected Calibration Error (ECE)** indicates systematic miscalibration on unrelated pairs. This confidence–accuracy inversion reflects a critical calibration failure observed in LLMs (Kadavath et al., 2022), with ECE values exceeding established thresholds for severe miscalibration (Guo et al., 2023), as shown in **Table 4**.

| **Category** | **ECE Range** | **Model Confidence Range** |
|---|---|---|
| Overall | ≈24.4-51.1% | ≈92-95% |
| Related pairs | ≈8.0-15.0% | ≈93–95% |
| Unrelated pairs | ≈24.0-51.0% | ≈91–94% |

**Table 4**: ECE for models on the Benchmark

We see degradation from 2–4x ECE increase in unrelated pairs, as illustrated in **Figure 2**. Overconfidence Error Rate (errors with confidence ≥0.75) ranges from 29.1–51.75%, meaning one-third to one-half of errors on unrelated pairs occur with high confidence, making them particularly dangerous in deployment contexts.

## 5.3 Semantic Collapse Rate

Semantic collapse rate (proportion of unrelated pairs misclassified as having relations) averages 37.6% across models, substantially lower than random guessing's expected 75% error. This indicates models do not fail through random guessing but through systematic generation of false relational structures (Liu et al., 2022), a known pathology in neural language models.

**Example:** When presented with an analogy such as *"Hospital is to flying as wolf is to_?"*, models often select an option by constructing a plausible relational narrative, for example invoking group membership or containment (e.g., wolves form packs), even though the base pair *hospital–flying* does not instantiate a meaningful semantic relation. The resulting explanation is internally coherent but grounded in a false premise.

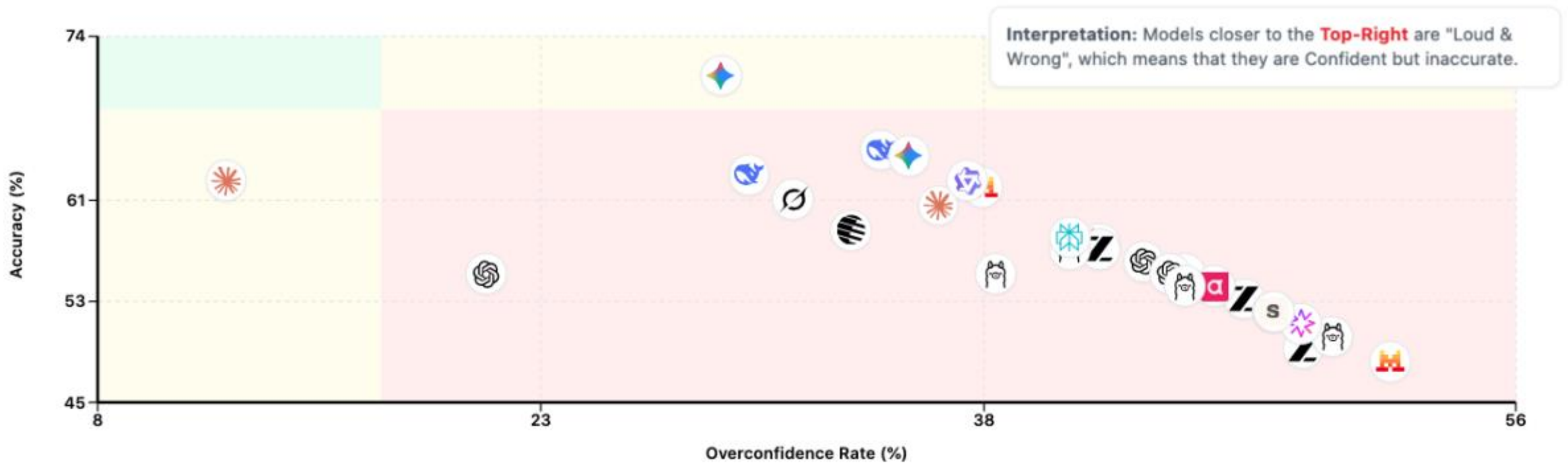

**Figure 3**: A comparison of models across Accuracy and Overconfidence Rate.

This behaviour illustrates a tendency to **fabricate relational structure** rather than explicitly recognize relation absence, suggesting that current models do not reliably represent unrelatedness as a distinct reasoning outcome.

### 5.4 Per-Relation Performance

On standard semantic relations, models achieve near-ceiling performance:

In contrast, as illustrated in **Figure 1**, unrelated pairs represent a distinct failure category. No model exceeds 41.35% accuracy, and several achieve 0%. This systematic pattern suggests a limitation in current modeling approaches, potentially arising from insufficient training signal or architectural constraints in representing the absence of semantic relations.

### 5.5 CORE Dataset Performance

On the 225K MCQ CORE dataset spanning 74 disciplines, accuracy drops to approximately 2%. This pronounced degradation indicates that the observed limitation extends to **domain-specific reasoning across diverse contexts**. This reflects findings that model performance degrades significantly when confronted with specialized reasoning across diverse contexts (Liang et al., 2022; Mialon et al., 2023).

### 5.6 Difficulty Stratification

Accuracy trends in **Table 5** indicate non-linear degradation: model performance improves from easy to medium questions but fails completely on hard questions, in contrast to the smoother decline observed for human accuracy. This is consistent with findings that sharp performance discontinuities often mask the absence of true underlying capability (Schaeffer et al., 2023).

| **Question Difficulty** | **Human Accuracy Range** | **Model Accuracy Range** |
|---|---|---|
| Easy | >90% | ≈52-71% |
| Medium | 70-90% | ≈72-86% |
| Hard | <70% | 0% |

**Table 5**: Accuracy by Question difficulty

## 6 Discussion

CORE isolates unrelatedness reasoning as a distinct and previously under-evaluated capability of LLMs. Despite strong performance on recognized relations, models consistently fail to identify absence of semantic relations while maintaining high confidence.

### 6.1 Universal Failure Across Models

The consistency of failures across **29 models** from different developers, parameter scales, and training paradigms suggests that the observed limitation is **not easily explained by variation in model size, developer, or standard training approach alone**. If the failure were primarily driven by idiosyncratic training data or optimization strategies, we would expect substantial variation across developers, model scales, or training regimes. Instead, we observe broadly similar behavior across these dimensions.

Specifically, unrelated-pair failures are consistently observed:

- **Across developers**, including OpenAI, Google, Anthropic, Meta, DeepSeek, and Mistral
- **Across model sizes**, ranging from approximately **8B to 405B parameters**
- **Across training approaches**, including supervised learning, RLHF, and reasoning-focused training

This cross-cutting consistency suggests that the failure reflects a **shared limitation in how current models and training pipelines handle relation absence**, potentially arising from a combination of architectural inductive biases, task formulation, and the availability of appropriate training signals. Prior work has similarly noted systematic generalization limits in transformer-based models across reasoning tasks (Hupkes et al., 2020; Rogers et al., 2020; Rytting & Wingate, 2021). While the common transformer backbone may contribute to this behavior, the results also indicate **substantial room for improvement through targeted data, objectives, and evaluation protocols** designed to explicitly model unrelatedness and uncertainty.

### 6.2 Architectural and Objective-Level Biases in Unrelatedness Reasoning

The observed failures likely arise from the interaction between **model architecture**, **training objectives**, and **evaluation formulation**, rather than from a single architectural limitation. Transformer-based models usually rely on attention mechanisms with softmax normalization, which distribute probability mass across candidate representations and do not naturally encode hard exclusion. While this does not preclude internal representations of uncertainty, it may bias models toward selecting and justifying one of the available options in closed-form reasoning and similar tasks.

This bias is particularly salient in **multiple-choice evaluation settings**, where models are required to select a single option even when the correct response corresponds to the absence of a meaningful semantic relation. From a relational completion perspective, such inputs are atypical: although one option correctly denotes unrelatedness, the formulation encourages models to search for and rationalize relational structure among competing alternatives, rather than explicitly reasoning about relation absence. As a result, models may prefer internally coherent but unsupported relational explanations over expressing uncertainty. This aligns with findings on model sycophancy, where models bias outputs to validate the user's implicit premises (Sharma et al., 2023).

Training objectives can further reinforce this behaviour. Standard cross-entropy loss rewards confident selection of correct answers but does not explicitly supervise uncertainty expression or abstention on ambiguous or ill-posed inputs. Consequently, models may learn to associate confidence with correctness in well-posed tasks, without acquiring complementary mechanisms for appropriately modulating confidence when no valid relation is present.

Taken together, these factors suggest a **systematic bias toward forced relational commitment** in unrelatedness scenarios. While this does not establish a definitive architectural cause, it highlights a mismatch between current modelling and training practices and the demands of reasoning about relation absence. Addressing this gap may benefit from targeted data, uncertainty-aware objectives, abstention mechanisms, and alternative evaluation formulations.

### 6.3 Confidence-Coherence Misalignment

A plausible contributing factor to the observed confidence-accuracy inversion is the role of **internal coherence** in model reasoning. When models generate explanations for relational decisions, even when those relations are spurious, the resulting explanations are often internally consistent and logically structured. For example, analogical reasoning such as "hospitals contain patients, flying involves aircraft, and wolves form packs" is coherent, despite being grounded in a false premise.

Prior work suggests that internal coherence and correctness are often correlated in standard reasoning settings, but that coherence alone does not guarantee faithful or correct reasoning (Jacovi & Goldberg, 2020; Turpin et al., 2023; Wiegreffe & Pinter, 2019). As a result, confidence estimates may become aligned with properties such as explanation consistency or structural plausibility rather than with factual or relational validity (Turpin et al., 2023; Zhao et al., 2024). When this correlation holds, confidence can serve as a useful proxy; however, in unrelatedness scenarios, coherence no longer tracks correctness.

Under this interpretation, models may assign high confidence to false but coherent explanations, leading to systematic miscalibration on unrelated pairs. In such cases, confidence reflects a proxy variable that correlates with correctness in well-posed tasks but fails when reasoning about relation absence. This hypothesis is consistent with the observed pattern of high confidence despite low accuracy, though establishing the underlying causal mechanisms remains an open direction for future work.

### 6.4 Implications

The combination of **high confidence and low accuracy** on unrelated pairs presents challenges for deploying language models in reasoning-dependent settings. When models confidently construct **spurious semantic relationships**, downstream systems may treat unsupported inferences as reliable signals.

In **healthcare**, such behaviour may surface high-confidence associations driven by confounding rather than causation, potentially influencing clinical decision-making. In **financial** contexts, models may assign undue significance to coincidental correlations, increasing exposure to risk. **Legal and scientific** applications face similar concerns, where plausible but incorrect relational reasoning may affect legal arguments or research prioritization.

Importantly, this failure mode differs from factual hallucination. Models do not invent entities; instead, they generate **internally coherent but incorrect relational structures**, which can be difficult to detect precisely because of their apparent plausibility.

#### 6.4.1 For Model Development

The findings highlight several directions for improving unrelatedness reasoning. Architectural approaches that treat relation absence as an explicit outcome, including alternative representations of negation or separation between relational inference and relation rejection, merit investigation. Training objectives may also be adapted to discourage confident errors on unrelated inputs, improve calibration on difficult cases, and encourage appropriate uncertainty on ill-posed tasks. Finally, unrelatedness reasoning and confidence alignment should be incorporated into optimization and evaluation objectives alongside standard accuracy.

#### 6.4.2 For Practitioners

Practitioners deploying LLMs in reasoning-dependent settings should consider additional safeguards. Models should be audited on benchmarks such as CORE prior to high-stakes deployment, with particular attention to performance on unrelated pairs. Confidence-based filtering can help flag potentially unreliable outputs, and cross-model agreement checks may identify cases requiring human review. Monitoring production outputs for patterns of semantic collapse can further reduce risk, especially in high-impact domains.

#### 6.4.3 For Evaluation and Benchmarking

These results suggest that unrelatedness reasoning should be treated as a standard evaluation dimension alongside existing benchmarks. Future work should extend such evaluations to domain-specific settings and track progress through shared benchmarks and leaderboards, enabling systematic assessment of proposed architectural and training interventions.

## 7 Work in Progress

Several important research directions extend this work:

**Multilingual Extension**: Preliminary multilingual experiments indicate even larger performance gaps in non-English languages, motivating extension of CORE to low-resource languages and broader multilingual evaluation.

**Fine-tuning Experiments**: Preliminary fine-tuning experiments on the full 225K-question CORE dataset show improvements in relational and general reasoning, which we plan to analyze and report in future work.

**Architectural Studies**: Mechanistic interpretability studies examining attention patterns, gradient flow, and linear activation directions associated with unanswerability (Lavi et al., 2025) could help identify contributing mechanisms. Testing of proposed architectural modifications (no-relation tokens, dual pathways, modified attention) would enable assessment of whether proposed solutions address root causes.

## 8 Limitations

Several limitations constrain the scope and interpretation of results:

**Language Scope**: All questions are in English. The findings may therefore reflect language-specific properties or relation saliency in English, and cross-lingual evaluation remains necessary for generalization claims.

**Format Limitations**: CORE evaluates multiple-choice format. Results may not directly transfer to open-ended generation where models must generate novel text. Multiple-choice provides explicit options that might scaffold performance differently than free-form generation.

**Text-Only Evaluation**: CORE is text-only. Multimodal reasoning with visual unrelated pairs is not tested. Results may not generalize to multimodal settings.

## Acknowledgments

We thank all participants who contributed to the human baseline evaluations and to the creation and validation of the CORE dataset and benchmark. We are also grateful to several academicians whose feedback helped shape this work. A list of contributors who consented to public acknowledgment is provided in Appendix C. We appreciate all contributions, including those not individually listed.

## Ethical Considerations

This work involves large-scale model inference, which entails significant computational cost and associated carbon emissions. We acknowledge this impact and encourage future research to adopt more efficient evaluation practices and transparent reporting of computational resources.

## Appendix A. Model Inference Prompt

```
CORE_PROMPT = """You will be given a multiple choice question with answer
options. Analyze the question carefully and provide your response in JSON
format only - no other text or explanation outside the JSON.

  **Question:**
  {question_text}

  {options_block}

  **Response Format:**
  Provide a single JSON object with exactly these fields:

  {{
    "answer": "A | B | C | D",
    "confidences_by_option": {{
      "A": <0-100>,
      "B": <0-100>,
      "C": <0-100>,
      "D": <0-100>
    }},
    "rationale": "brief explanation of reasoning",
    "time_to_think": <integer_seconds>,
    "difficulty_rating": <integer_1_to_5>,
    "novelty": "seen | unseen",
    "hallucinating": <true | false>,
    "ambiguous_question": <true | false>,
    "reasoning_type": "factual recall | logical deduction | elimination |
guessing | inferential reasoning | pattern recognition | contextual
understanding",
    "supporting_facts": "evidence supporting answer",
    "confidence_type": "prior knowledge | strong elimination | partial match
| intuition",
    "token_count_to_answer": <integer_token_count>,
    "was_revised": <true | false>,
    "would_you_stake_on_it": <true | false>,
    "uncertainty_expressed": <true | false>,
    "user_needs_verification": <true | false>
  }}

  **Field Specifications:**
  - `answer`: Selected option letter (A/B/C/D/etc.)
  - `confidences_by_option`: Confidence score 0-100 for each option
(independent scores, don't need to sum to 100)
  - `rationale`: Brief explanation (1-2 sentences)
  - `time_to_think`: Estimated seconds spent reasoning (integer)
  - `difficulty_rating`: 1 (very easy) to 5 (very hard)
  - `novelty`: "seen" or "unseen"
  - `hallucinating`: true if uncertain/confabulating, false otherwise
  - `ambiguous_question`: true if question is unclear, false otherwise
  - `reasoning_type`: One of: "factual recall", "logical deduction",
"elimination", "guessing", "inferential reasoning", "pattern recognition",
"contextual understanding"
  - `supporting_facts`: Evidence or reasoning that supports your answer
  - `confidence_type`: One of: "prior knowledge", "strong elimination",
"partial match", "intuition"
  - `token_count_to_answer`: Estimated tokens used internally (integer)
```

```
  - `was_revised`: true if you changed your initial answer, false otherwise
  - `would_you_stake_on_it`: true if confident enough for high-stakes use,
false otherwise
  - `uncertainty_expressed`: true if rationale shows uncertainty, false
otherwise
  - `user_needs_verification`: true if human verification recommended, false
otherwise

Return ONLY valid JSON."""
```

**Note on Model Introspection:** We acknowledge that LLMs lack access to internal system states and cannot reliably report metrics such as time_to_think, token_count_to_answer, or hallucinating. These fields are collected not as ground-truth data, but to analyse **metacognitive calibration** and **simulation behavior**, specifically, to determine if the model's self-reported effort and uncertainty correlate with empirical accuracy or if they represent hallucinated "introspective illusions." This analysis extends to was_revised and rationale, checking for post-hoc rationalizations where the model invents a narrative to justify a selected answer.

## Appendix B. Metrics covering various aspects of Model Performance

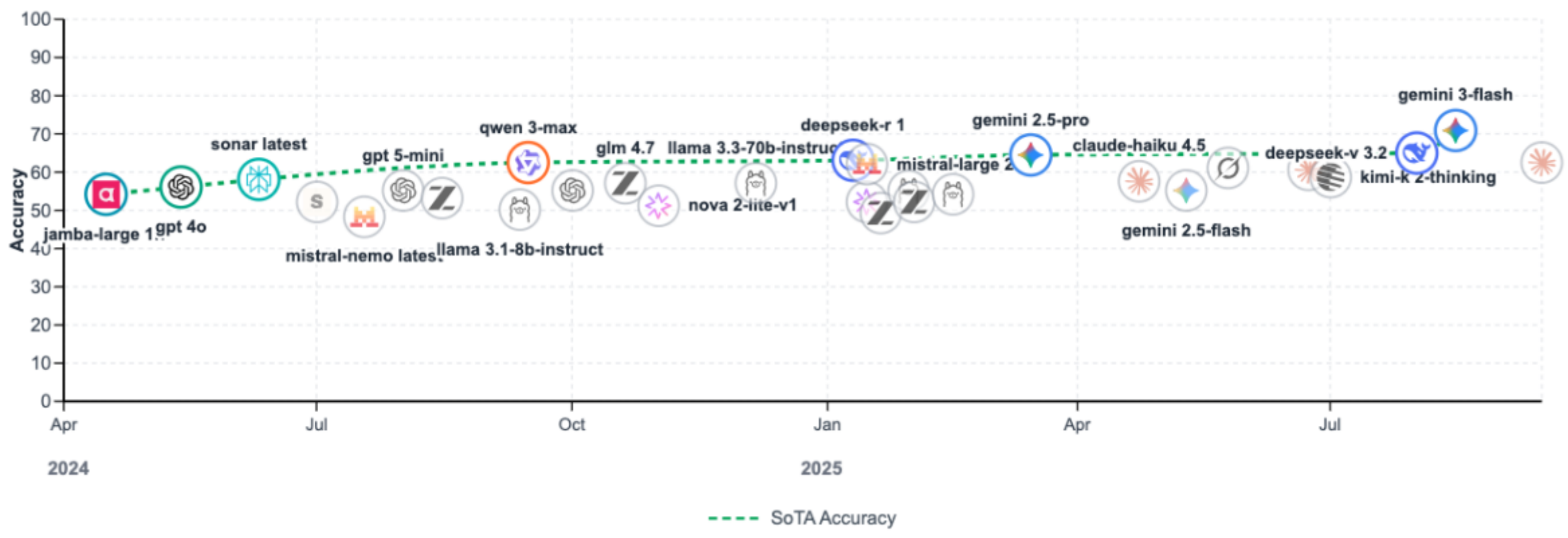

**Figure 4**: Accuracy trends across evaluated models.

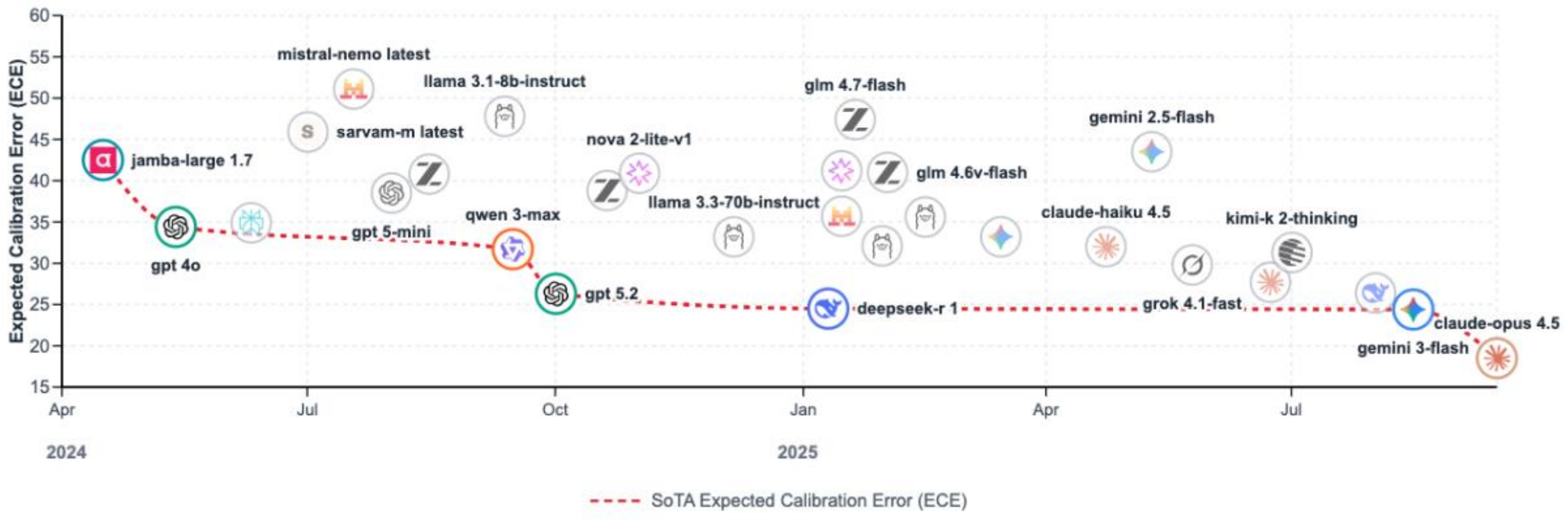

**Figure 5**: ECE trends across evaluated models.

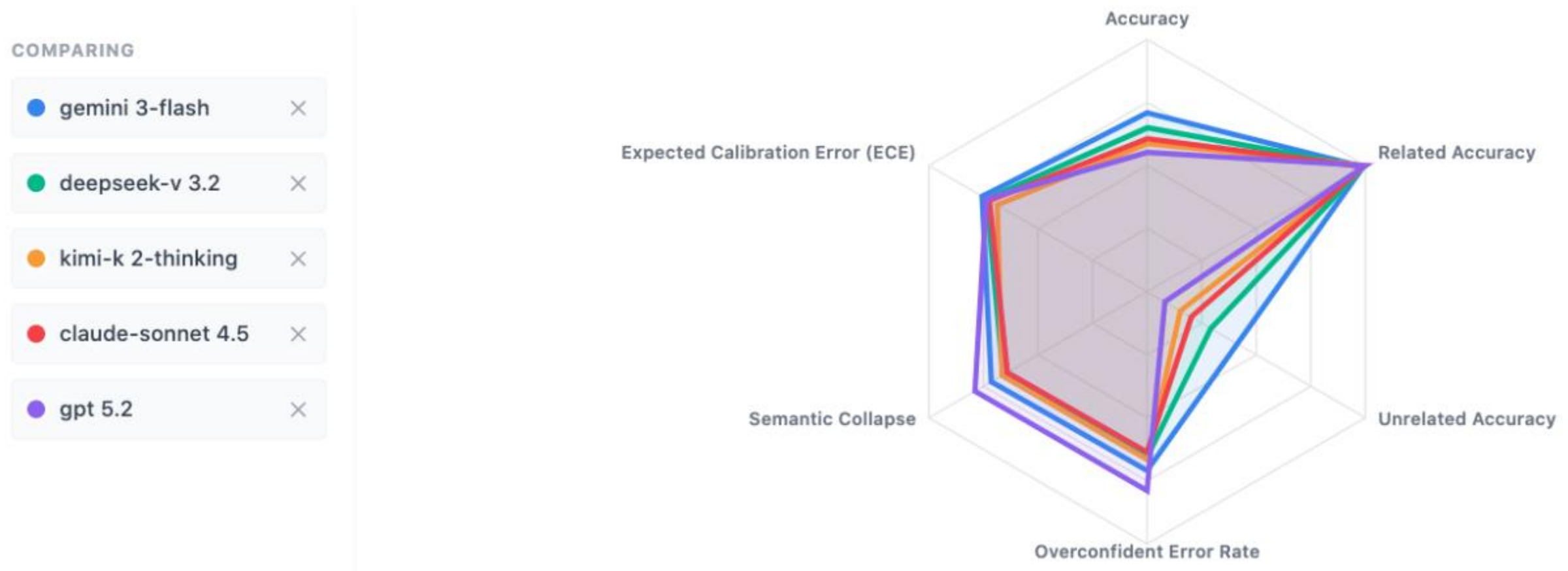

**Figure 6**: A comparison of top models from different developers across critical metrics.

## Appendix C. Team

### Organizing team

Satyam Dwivedi[1], Sanjukta Ghosh[5], Shivam Dwivedi[5], Nishi Kumari[1], Anil Kumar Thakur[5], Anurag Purushottam[1], Deepak Alok[6], Praveen Gatla[2], Manjuprasad B[4], Bipasha Patgiri[7]

### Human Baseline Contributors

This list includes only the names of participants who consented to public acknowledgment of their contributions. We are equally grateful to all participants who took part in this effort.

Aman Gupta [2], Anjali Kumari [2], Ankit Raj Gupta [2], Ankita Keshri[2], Bhaskar Singh[2], Bipasha Paul[2], Chitranshi Tiwari [2], Deepanshu Patel[2], Harsh Mishra [2], Himesh Jee Amar[2], Kajal Kumari[2], Mahi Doshi [2], Muskan Chaudhary [2], Nancy Mittal[2], Priyanshu Kumar [2], Rahul Kumar[2], Rameesa Azma[2], Rasi Shil[2], Vineet Kumar [2], Warisha Quatil [2], Subhash Bharti[3], Achala C[4], Ananya R Naik[4], Ananyabm[4], Anjali Ajith [4], Ankitha Ks[4], Ashitha [4], Ayesha Banu[4], B.Tanmayi[4], Basavasiri H L [4], Bhagyashree Hokrani [4], Bhoomika.P[4], Chinmayi Mohan [4], Dhanalakshmi.N[4], Divya Kn[4], E.Sai Sruthi[4], Hanasi Matada Eshwari [4], Harini Nayaka Gm [4], Harshitha R[4], Jeevika Ks [4], Keerthana B[4], Lisha S Kumar [4], M.R.Meghana[4], Manogna Keshav[4], Manya. E. A[4], Meghana M[4], Megharani[4], Mohammed Ayesha Tahreem[4], Nanditha M [4], Nithyashree.H [4], Punyashree P R[4], R Veena[4], Rakshitha M[4], Ruchitha S[4], Sahana.D.S[4], Sai Pallavi [4], Sameeksha S[4], Sandhya R [4], Sanika [4], Sanjana G Rao [4], Shafna Ms [4], Sharanya S Prasad [4], Shravya.H[4], Shruthi Reddy[4], Sinchana[4], Sinchana S[4], Siri N Murthy[4], Siri Patel M [4], Sk Nikitha Reddy [4], Snehaganga N S[4], Sowndarya B[4], Spoorthi U[4], Subhangi Dutta[4], Syeda Saneen [4], Tammisetty Harini[4], Thanushree M R [4], Thanushree S T [4], Varsha Suresh[4], Varshitha S [4], A Vijay Aditya[5], Aasish[5], Aayush Bhat[5], Abhijeet Singh[5], Abhishek Chauhan[5], Abhishek Kumar Maurya[5], Abhyudaya[5], Adarsh Kumar Gupta [5], Addagalla Lakshmi Sowjanya[5], Adi Akhilesh Singh[5], Aditi Gupta[5], Aditya Prakash [5], Aditya R Jadhav[5], Aditya Raj[5], Aditya Singh[5], Aishwarya Agnihotri[5], Ajay Patel[5], Ajay Singh[5], Akanksha Singh[5], Akshita Ravichndran[5], Akula Manasa [5], Allu Deekshita[5], Aman Kumar[5], Aman Kumar Yadav[5], Amardeep Jarwal[5], Amit Negi [5], Amit Singh [5], Angraj Shah[5], Anjali Kumari[5], Anjaneya Raj Garg[5], Ankit Prakash[5], Ankit Sinha[5], Anshu Kumar Ram[5], Anshu Yadav[5], Anupurba Dhara[5], Anurag Kamboj[5], Anushka Choudhary[5], Arushi Gupta[5], Aryan[5], Aryan Parihar[5], Ashutosh Singh[5], Atmadeep Bhattacharya[5], Avanish Dhapare[5], Awaneesh Kumar Pandey[5], Ayush Barot [5], Ayush Kumar[5], Ayush Mondal[5], Ayush Sharma[5], Ayush Tripathi[5], Banoth Nandineshwar[5], Bellala Mukesh[5], Bhanu Verma[5], Bhavya Singh [5], Bhupendra Yadav[5], Bommadi Mukesh Kumar Reddy[5], Brijesh Kumar[5], Chaudhary Digvijay Daniel Singh[5], Check__Email[5], Chelsi Narang[5], Chennadi Pavan Sainath Reddy[5], Chivukula Sri Eswar Balaji[5], Deen Dayal Prajapati[5], Deepaprakash K[5], Deepjyoti Rabha[5], Dhruvi Rajeshbhai Mahyavanshi[5], Dipti Gupta[5], Divyanshu Yadav[5], Durgam Arun Kumar[5], Faiz Aman[5], Farah Adiba[5], Fizaan Khan[5], Ganesh Sakkarge[5], Ganguly Singh[5], Gaurish Maheshwari[5], H Poojan[5], Happy Kannaujiya[5], Harsh[5], Harsh Kadiyan[5], Harsh Kumar[5], Harsh Vardhan[5], Harshit Virmani[5], Harshita Rajput[5], Harshvardhan Goplani[5], Hrishabh Deshmukh[5], Ishika Saini[5], Jagat Jyoti Sarkar[5], Jain Aditya Avinash[5], Jayesh Sewlani[5], Kali Chopra[5], Kalyanam Pranay[5], Kamal[5], Kanukollu Sateesh Kumar [5], Karishma Santani[5], Kartikeya Pandey[5], Kaushik Kumar[5], Kishore Nayak D[5], Kolgane Sanskruti Sanjay[5], Komal Bhalotia[5], Kratika Maheshwari[5], Kritarth[5], Kshitij Kumar[5], Kumar Pundareekaksh[5], Kumar Shubham[5], Kushagr Kapoor[5], Lalit Tolani[5], Lalithya C [5], M Balasubramanian[5], Madhur Vilas Bahadure[5], Malipatel Sravan Kumar Reddy[5], Manas Jayaswal[5], Manav Gangwar [5], Manish Kumar[5], Manisha Bishnoi[5], Mannuru Venkateswarlu[5], Mayank Agrawal [5], Moksh[5], Motilal Bhatra[5], Mradul Misra[5], Mridul Gupta[5], Mrinal Jain[5], Naitik Jain[5], Nakshatra Shivhare [5], Neelam Meena[5], Nida Rahma.A[5], Nikhil Deshmukh[5], Nikita Gupta[5], Nimish Thakur [5], Nitesh Singh[5], Nitin Agrawal[5], Nitish Kumar[5], Ojasvi Tripathi[5], Ojaswi Pandey[5], Om Abhishek[5], Om Shankar[5], Omkaran [5], Pakhi Awasthi [5], Parteek Yadav[5], Pavani Gupta [5], Pooja Yadav[5], Prabhankur[5], Prafull Kumar Deepak[5], Prakhar Nema[5], Pranjali Yadav[5], Prashant Nautiyal[5], Prathu Tripathi [5], Priti Kumari[5], Priyadarshi Annupam[5], Priyanshu Tiwari[5], Purnima Singh[5], Rachit Mittal[5], Ragula Eeshareddy[5], Rahul Kumar Sonkar[5], Rajat Varshney[5], Ramgopal Verma[5], Raskar Aniket Dattatray[5], Ratnesh Kumar Sharma [5], Ravi Kumar[5], Rishabh Singh[5],

Rishabh Yadav[5], Rishi Mishra[5], Rishi Soni[5], Rishit Pal[5], Ritesh Soni[5], Ritik Rai[5], Ritik Raj [5], Ritik Raushan[5], Rituraj Barai[5], Rohan Sharma[5], Rohit Pandey[5], Rohit Prasad[5], Saarang Kumar[5], Sagar Sachan[5], Sahil Shekhar[5], Saksham Goel[5], Samarth Jain[5], Samir Kumar [5], Sammit Dhar[5], Sampat Meena[5], Sapavat Sravan[5], Saptarshi Chakraborty[5], Sarthak Shewale[5], Saurabh Kumar[5], Shashank Kumar[5], Sheetal Nagar[5], Shikha Kaloniya [5], Shivangi Gupta[5], Shivansh Gupta[5], Shivanshu Kumar[5], Shreyam Chaurasia [5], Shreyansh Singh[5], Shrija Tiwary[5], Shubham Kumar[5], Shubham Patel[5], Shubhendra Taneja[5], Siddhant Bhardwaj[5], Siddharth Prakash[5], Soham Abhay Kadam[5], Sonali Singh[5], Sonu Sourabh[5], Sourashis Das[5], Soustab Haldar[5], Sparsh Gupta [5], Srajan Seth[5], Srishti Jaiswal[5], Sudhanshu Ranjan[5], Suharsh Sonkar [5], Suman Kumar[5], Sunanda Pandey[5], Surkanti Harshitha Reddy[5], Sushank [5], Swapnil Wakankar [5], Tanay Ahir[5], Tanish Jangir [5], Tanishka Nama[5], Tanishq Gupta[5], Tarani Mishra[5], Tejavath Sudhakar[5], Tushar Sarda[5], Udeechi Srivastav[5], Utkarsh Srivastava[5], Vaddadi Lakshmi Sri Sai Srinivas[5], Vadithya Rajagopal[5], Vaibhav Jain[5], Vaibhav Saini[5], Vanshika[5], Vedant Bhoruka[5], Vijay Kumar [5], Vikash Kumar[5], Vineet Tyagi[5], Vipul Bharti[5], Vishal[5], Vishisht Dubey[5], Vishnu Katara[5], Vishvender Pachaar[5], Vivek Kumar[5], Yash Agarwal[5], Yash Sachan[5], Aadithya Balachandran[6], Abhay[6], Aditya Sharma[6], Akshay[6], Barza A K[6], Bhishen Kumar Sahu[6], Chinmayee Mohapatra[6], Dhairya Yadav[6], Divyanka Swarna[6], Hariprasad Doley[6], Karthika P[6], Khushi[6], Lugai Kamei[6], Manasi Anil Lamsoge[6], Mojum Kamduk[6], Neeraj N Shetty[6], Panduru Tanisha[6], Rohit B Sharma[6], Sai Sudeep Das[6], Sara Singh[6], Sharon Valui[6], Sheersha Roy[6], Shivang Jaiswal [6], Shweta Umrankar[6], Soumya Jain[6], Sumayya Ayesha[6], Suvrojit Nath[6], Tanisha[6], Vanshika Gupta[6], Zitaksyor Sonowal[6], Mohima Narzary[7], Pratiksha Rabha[7], Ruba Das[7], Shruti Dekaraja[7], Yuktashree Hazarika[7]

**Affiliations**

[1]Vaikhari AI, [2]BHU, [3]Galgotias University, [4]GSSSIETW, [5]IIT BHU, [6]IIT Delhi, [7]Tezpur University